\documentclass[runningheads]{llncs}
\usepackage{graphicx}
\usepackage{amsmath,graphicx}
\usepackage{times}
\usepackage[dvipsnames]{xcolor}

\usepackage{tabularx,booktabs}
\usepackage{arydshln}
\usepackage{tabu}
\usepackage{adjustbox}
\usepackage{enumerate,multicol}
\usepackage{multirow}
\usepackage{multicol}
\usepackage{enumitem}
\usepackage{amsmath,array,graphicx}
\usepackage{gensymb}
\usepackage{float}
\usepackage{caption}
\usepackage{subcaption}
\usepackage{balance}
\usepackage{xspace}
\usepackage{amssymb}
\usepackage{pifont}
\usepackage{subcaption}
\usepackage{tabularx}
\usepackage{amsmath}
\usepackage{array}
\newcommand{\xmark}{\ding{55}}

\usepackage{times}

\usepackage{soul}
\usepackage{url}
\usepackage[utf8]{inputenc}
\usepackage{amsmath}
\usepackage{booktabs}

\usepackage{hyperref}

\hypersetup{
    colorlinks,
    linkcolor={red!50!black},
    citecolor={blue!50!black},
    urlcolor={blue}
}

\newif\ifdraft
\drafttrue

\definecolor{orange}{rgb}{1,0.5,0}
\definecolor{pink}{rgb}{0.98, 0.38, 0.5}
\definecolor{darkgreen}{rgb}{0.055, 0.490, 0.016} 

\ifdraft
 \usepackage[normalem]{ulem}
 \newcommand{\RS}[1]{{\color{red}{\bf RS: #1}}}
 
 \newcommand{\PMN}[1]{{\color{orange}{\bf PMN: #1}}}
 
 \newcommand{\JGT}[1]{{\color{blue} JGT: #1}}

\else
 \newcommand{\sout}[1]{}
 \newcommand{\RS}[1]{{\color{red}{}}}
 
 \newcommand{\PMN}[1]{{\color{red}{}}}
 
 \newcommand{\JGT}[1]{{\color{blue}{}}}
 
\fi

\newcolumntype{S}{>{\centering\arraybackslash}p{1.05cm}}
\newcolumntype{L}{>{\small}l}
\newcolumntype{T}{>{\centering\arraybackslash}p{1.1cm}}

\newcommand{\comment}[1]{}

\begin{document}
\title{DeepPyramid: Enabling Pyramid View and Deformable Pyramid Reception for Semantic Segmentation in Cataract Surgery Videos\thanks{This work was funded by Haag-Streit Switzerland and the FWF Austrian Science Fund under grant P 31486-N31.}}

\titlerunning{DeepPyramid: Pyramid View and Deformable Pyramid Reception}

\author{Negin Ghamsarian\inst{1} 
\and Mario Taschwer\inst{2}
\and Raphael Sznitman \inst{1} 
\and Klaus Schoeffmann \inst{2} 
} 

\institute{Center for AI in Medicine, Faculty of Medicine, University of Bern\\ 
\email{\{negin.ghamsarian, raphael.sznitman\}@unibe.ch}
\and Department of Information Technology, Alpen-Adria-Universit\"at Klagenfurt \email{\{mt,ks\}@itec.aau.at}
}

\authorrunning{N. Ghamsarian et al.}
\maketitle              

\begin{abstract}
Semantic segmentation in cataract surgery has a wide range of applications contributing to surgical outcome enhancement and clinical risk reduction. However, the varying issues in segmenting the different relevant structures in these surgeries make the designation of a unique network quite challenging. This paper proposes a semantic segmentation network, termed DeepPyramid, that can deal with these challenges using three novelties: (1) a Pyramid View Fusion module which provides a varying-angle global view of the surrounding region centering at each pixel position in the input convolutional feature map; (2) a Deformable Pyramid Reception module which enables a wide deformable receptive field that can adapt to geometric transformations in the object of interest; and (3) a dedicated Pyramid Loss that adaptively supervises multi-scale semantic feature maps. Combined, we show that these modules can effectively boost semantic segmentation performance, especially in the case of transparency, deformability, scalability, and blunt edges in objects. We demonstrate that our approach performs at a state-of-the-art level and outperforms a number of existing methods with a large margin ($3.66\%$ overall improvement in intersection over union compared to the best rival approach). 

\end{abstract}

\section{Introduction}
\label{sec:introduction}

Cataracts are naturally developing opacity that obfuscates sight and is the leading cause of blindness worldwide, with over 100 million people suffering from them. Today, surgery is the most effective way to cure patients by replacing natural eye lenses with artificial ones. More than 10 million cataract surgeries are performed every year, making it one of the most common surgeries globally~\cite{JFCS}. With the aging world population growing, the number of patients at risk of complete cataract-caused blindness is sharply increasing~\cite{Wang2017} and the number of surgeries needed brings unprecedented organizational and logistical challenges.

\begin{figure}[t]
    \centering
    \includegraphics[width=0.99\columnwidth]{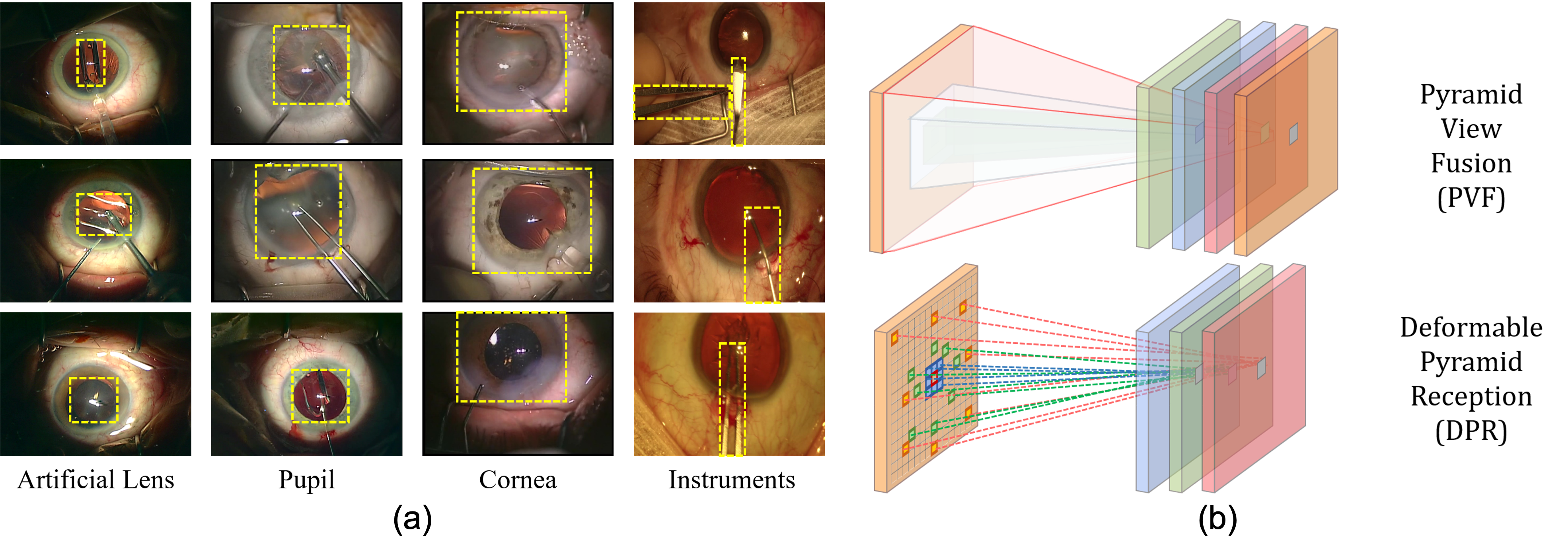}
    \caption{(a) Challenges in semantic segmentation of relevant objects in cataract surgery. (b) Proposed Pyramid View Fusion and Deformable Pyramid Reception modules. }
    \label{fig: Problems}
\end{figure}

To help train future surgeons and optimize surgical workflows, automated methods that analyze cataract surgery videos have gained significant traction in the last decade. With the prospect of reducing intra-operative and post-operative complications~\cite{RBE}, recent methods have included surgical skill assessment~\cite{RDC,DeepPhase}, remaining surgical time estimation~\cite{Marafioti21}, irregularity detection~\cite{LensID} or relevance-based compression~\cite{RelComp}. In addition, a reliable relevant-instance-segmentation approach is often a prerequisite for a majority of these applications~\cite{Pissas21}. In this regard, four different structures are typically of interest: the intraocular lens, the pupil, the cornea, and surgical instruments. Due to the diversity in the appearance of these structures, segmentation methods must overcome several hurdles to perform well on real-world video sequences. Specifically, a semantic segmentation network is required to simultaneously deal with: 1) a transparent artificial lens that undergoes deformations, 2) color, shape, size, and texture variations in the pupil, 3) unclear edges of the cornea, and 4) severe motion blur, reflection distortion, and scale variations in instruments (see Fig.~\ref{fig: Problems}-a). This work looks to provide a method to segment these structures despite the mentioned challenges. 

Several network architectures for cataract surgery semantic segmentation have been proposed or have been used in the recent past~\cite{RAUNet,PAANet,BARNet,DeepLabv3+,PSPNet}. Many of these methods have been based on the U-Net architecture~\cite{U-Net} and aimed at improving accuracy by addressing different limitations from the original architecture. In~\cite{RAUNet,PAANet}, different attention modules were used to guide the network's computational efforts toward the most discriminative features in the input feature map considering the characteristics of the objects of interest. Additionally, fusion modules have been proposed to improve semantic representation via combining several feature maps~\cite{DeepLabv3+,PSPNet}. However, as we show in our experiments, these methods still have difficulties with the aforementioned challenges in cataract video segmentation. 

In this work, we propose a novel architecture that is tailored to adaptively capture semantic information despite the challenges typically found in cataract surgery videos. Our proposed network, {\it DeepPyramid}\footnote[1]{The PyTorch implementation of DeepPyramid is publicly available at \url{https://github.com/Negin-Ghamsarian/DeepPyramid_MICCAI2022}}, introduces three key contributions: (i) a Pyramid View Fusion (PVF) module allowing a varying-angle surrounding view of the feature maps for each pixel position, (ii) a Deformable Pyramid Reception (DPR) module, which enables a large, sparse, and learnable receptive field to perform shape-wise feature extraction (see Fig.~\ref{fig: Problems}-b), and (iii) a Pyramid Loss, ($P\mathcal{L}$) to explicitly supervise multi-scale semantic feature maps in our network. We show in the experiments that our approach outperforms by a significant margin twelve rival state-of-the-art approaches for cataract surgery segmentation. Specifically, we show that our model is particularly effective for deformable, transparent, and changing scale objects. In addition, we show the contribution of each of the proposed additions, highlighting that the addition of all three yields the observed improvements.

\section{Methodology}
\label{sec: Methodology}
\begin{figure*}[t]
    \centering
    \includegraphics[width=0.99\textwidth]{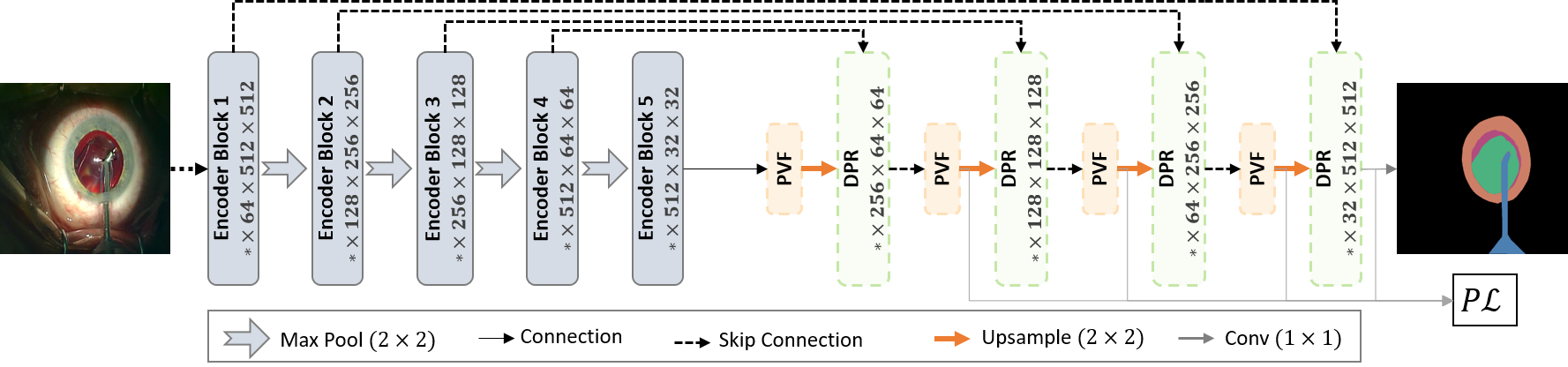}
    \caption{Overall architecture of the proposed DeepPyramid network. It contains Pyramid View Fusion (PVF), Deformable Pyramid Reception (DPR), and Pyramid Loss ($P\mathcal{L}$).}
    \label{fig: Block_diagram}
\end{figure*}
Our proposed segmentation strategy aims to explicitly model deformations and context within its architecture. Using a U-Net-based architecture, our proposed model is illustrated in Fig.~\ref{fig: Block_diagram}. At its core, the encoder network remains that of a standard VGG16 network. Our approach is to provide useful decoder modules to help alleviate segmentation concerning relevant objects' features in cataract surgery \footnote[2]{Since changing the encoder network entails pretraining on a large dataset (such as ImageNet), which in turn imposes more computational costs, we only add the proposed modules after the bottleneck. Nevertheless, since these modules are applied to concatenated features coming from the encoder network via skip connections, the encoder features can be effectively guided.}. Specifically, we propose a Pyramid View Fusion (PVF) module and a Deformable Pyramid Reception (DPR) module. These are then trained using a dedicated Pyramid Loss ($P\mathcal{L}$). 

Conceptually, the PVF module is inspired by the human visual system and aims to recognize semantic information found in images considering not only the internal object's content but also the relative information between the object and its surrounding area. Thus the role of the PVF is to reinforce the observation of relative information at every distinct pixel position. Specifically, we use average pooling to fuse the multi-angle local information for this novel attention mechanism. Conversely, our DPR module hinges on a novel deformable block based on dilated convolutions that can help recognize each pixel position's semantic label based on its cross-dependencies with varying-distance surrounding pixels without imposing additional trainable parameters. Due to the inflexible rectangle shape of the receptive field in regular convolutional layers, the feature extraction procedure cannot be adapted to complex deformable shapes \cite{DefED-Net}. Our proposed dilated deformable convolutional layers attempt to remedy this explicitly in terms of both scale and shape. We now specify these modules and our loss function in the following subsections. 

\begin{figure*}[t]
    \centering
    \includegraphics[width=0.99\textwidth]{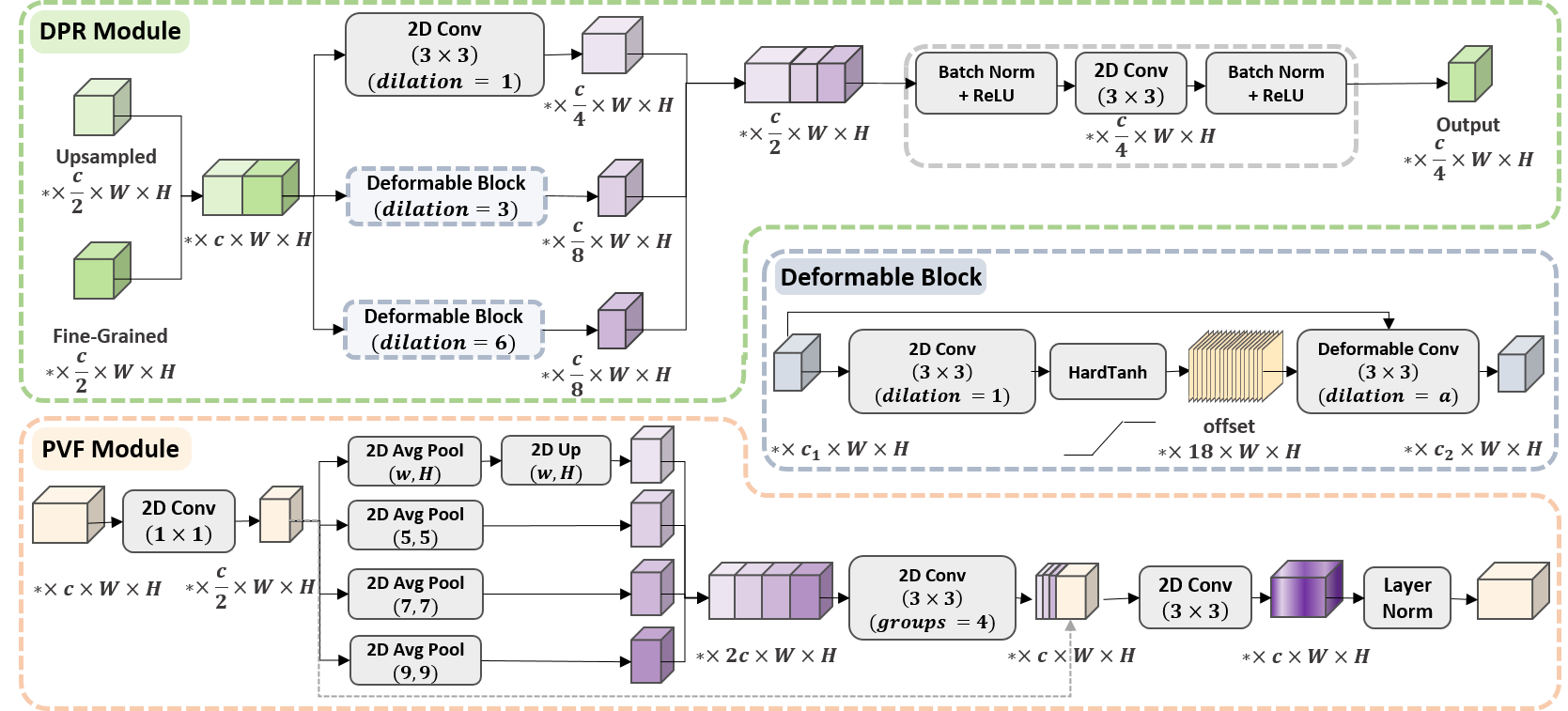}
    \caption{Detailed architecture of the Deformable Pyramid Reception (DPR) and Pyramid View Fusion (PVF) modules.}
    \label{fig: DPR}
\end{figure*}

\paragraph{\textbf{Pyramid View Fusion (PVF).}} 
First, a bottleneck is formed by employing a convolutional layer with a kernel size of one to curb computational complexity. The convolutional feature map is then fed into four parallel branches: a global average pooling layer followed by upsampling and three average pooling layers with progressively larger filter sizes and a common stride of 1. Note that using a one-pixel stride is essential to obtain pixel-wise centralized pyramid views in contrast with region-wise pyramid attention as shown in PSPNet~\cite{PSPNet}. The output feature maps are then concatenated and fed into a convolutional layer with four groups. This layer is responsible for extracting inter-channel dependencies during dimensionality reduction. A regular convolutional layer is then applied to extract joint intra-channel and inter-channel dependencies before being fed into a layer-normalization function. A summary of this module is illustrated in Fig.~\ref{fig: DPR}.

\paragraph{\textbf{Deformable Pyramid Reception (DPR).}}
As shown in Fig.~\ref{fig: DPR} (top), the fine-grained feature map from the encoder and coarse-grained semantic feature map from the previous layer are first concatenated. These features are then fed into three parallel branches: a regular $3\times 3$ convolution and two deformable blocks with different dilation rates. Together, these layers cover a learnable but sparse receptive field of size $15\times 15$ \footnote[3]{The structured $3\times 3$ filter covers up to 1 pixel from the central pixel. The deformable filter with $dilation=3$ covers an area of 2 to 4 pixels away from each central pixel. Similarly, the deformable convolution with $dilation=6$ covers an area of 5 to 7 pixels away from each central pixel. Together, these form a sparse filter of size $15\times 15$ pixels.} as shown in~Fig.~\ref{fig: Problems}~(b). The output feature maps are then concatenated before undergoing a sequence of regular layers for higher-order feature extraction and dimensionality reduction.

The deformable blocks used in the DPR module consist of a regular convolutional layer applied to the input feature map to compute an offset field for deformable convolution. The offset field provides two values per element in the convolutional filter (horizontal and vertical offsets). Accordingly, the number of offset field's output channels for a kernel of size $3\times 3$ is equal to 18. 
Inspired by dU-Net~\cite{dU-Net}, the convolutional layer for the offset field is followed by an activation function, which we set to the hard tangent hyperbolic function, as it is computationally efficient and clips offset values to the range of $[-1,1]$. In summary (see Fig.~\ref{fig: DPR}), the deformable block uses learned offset values along with the convolutional feature map with a predetermined dilation rate to extract object-adaptive features.

The output feature map ($y$) for each pixel position ($p_0$) and the receptive field ($\mathcal{RF}$) for a regular 2D convolution with a $3\times 3$ filter and dilation rate of 1 can be computed by, 
\begin{equation}
y(p_o)=\sum_{p_i\in \mathcal{RF}_1}{x(p_0+p_i)w(p_i)},
\end{equation}
where $\mathcal{RF}_1=\{(-1,-1),(-1,0), ...,$ $(1,0),(1,1)\}$, $x$ denotes the input convolutional feature map, and $w$ refers to the weights of the convolutional kernel. In a dilated 2D convolution with a dilation rate of $\alpha$, the receptive field can be defined as $\mathcal{RF}_{\alpha}=\alpha \times \mathcal{RF}_1$. Although the sampling locations in a dilated receptive field have a greater distance to the central pixel, they follow a firm structure. In a deformable dilated convolution with a dilation rate of $\alpha$, the sampling locations of the receptive field are dependent on the local contextual features. In the proposed deformable block, the sampling location for the $i$th element of the receptive field and the input pixel $p_0$ are calculated as,
\begin{equation}
   \mathcal{RF}_{def,\alpha}[i,p_{0}]= \mathcal{RF}_\alpha[i]
   +f\left(\sum_{p_j\in \mathcal{RF}_1}{x(p_0+p_j)\hat{w}(p_j)}\right),
\label{eq: RF3}
\end{equation}
\noindent
where $f$ denotes the activation function, which is the tangent hyperbolic function in our case, and $\hat{w}$ refers to the weights of the offset filter. This learnable receptive field can be adapted to every distinct pixel in the convolutional feature map and allows the convolutional layer to extract stronger informative semantic features when compared to the regular convolution.

\paragraph{\textbf{Pyramid Loss (P$\mathcal{L}$).}}
To train our network using the PVF and DPR modules, we wish to directly supervise the multi-scale semantic feature maps of the decoder. To enable direct supervision, a depth-wise fully connected layer is formed using a pixel-wise convolution operation. The output feature map presents the semantic segmentation results with the same resolution as the input feature map. To compute the loss for varying-scale outputs, we downscale the ground-truth masks using inter-nearest downsampling for multi-class segmentation and max-pooling for binary segmentation. Our overall loss is then defined as, 
\begin{equation}
P\mathcal{L} = \mathcal{L}_1 + \alpha \mathcal{L}_2 + \beta \mathcal{L}_4 + \gamma \mathcal{L}_8,
\label{eq: pl_weights}
\end{equation}
where $\alpha$, $\beta$, and $\gamma$ are predetermined weights in the range of $[0,1]$ and $\mathcal{L}_i$ denotes the loss of output mask segmentation result with the resolution of $(1/i)$ compared to the input resolution.

\section{Experimental Setup}
\label{sec: Experimental Setup}
To evaluate the performance of our approach, we make use of data from three datasets. These include the ``Cornea''~\cite{RelComp} and ``Instruments" mask annotations from the CaDIS dataset~\cite{CaDIS}. In addition, we have collected a separate dataset from which we performed the ``Intraocular Lens" and ``Pupil" pixel-wise segmentations\footnote[4]{The customized datasets is publicly released in \url{https://ftp.itec.aau.at/datasets/ovid/DeepPyram/}.}. The total number of training and test images for the aforementioned objects are 178:84, 3190:459, 141:48, and 141:48, respectively\footnote[5]{Our evaluations are based on binary segmentation per relevant object so that we do not have the imbalance problem. In the case of multi-class classification, methods such as oversampling can mitigate the imbalance problem \cite{ESSiCS}.}. In the following experiments, all training and test images were split patient-wise to ensure realistic conditions. 

We compare the performance of DeepPyramid with thirteen different state-of-the-art segmentation approaches including UNet++ and UNet++/DS~\cite{UNet++}, CPFNet \cite{CPFNet}, BARNet~\cite{BARNet}, PAANet \cite{PAANet}, dU-Net\footnote[6]{Our version of du-Net has the same number of filter-response maps as the U-Net.}~\cite{dU-Net}, MultiResUNet~\cite{MultiResUNet}, CE-Net \cite{CE-Net}, RAUNet~\cite{RAUNet}, FED-Net~\cite{FED-Net}\, UPerNet \cite{UPerNet}, PSPNet\footnote[7]{To provide a fair comparison, we adopt our improved version of PSPNet, featuring a decoder designed similarly to U-Net (with four sequences of double-convolution blocks).} \cite{PSPNet}, and U-Net~\cite{U-Net}\footnote[8]{BARNet, PAANet, and RAUNet are tailored for instrument segmentation in surgical videos. Other methods are state-of-the-art for medical image segmentation.}. With the exception of the U-Net, MultiResUNet, and dU-Net, which do not use a pretrained backbone, the weights of the backbone for all networks were initialized with ImageNet~\cite{ImageNet} training weights.
The input size of all models is set to $3\times 512\times 512$. 

For all methods, training is performed using data augmentation. Transformations considered the inherent and statistical features of datasets. For instance, we use motion blur transformation to encourage the network to deal with harsh motion blur regularly occurring in cataract surgery videos. We further use brightness and contrast, shift and scale, and rotate augmentation. 

Due to the different depth and connections of the proposed and rival approaches, all networks are trained with three different initial learning rates ($lr\in\{5,2,10\}\times 10^{-4}$), and the results with the highest IoU for each network are listed. The learning rate is scheduled to decrease every two epochs with the factor of $0.8$. In all evaluations, the networks are trained end-to-end and for 30 epochs. We use a threshold of $0.1$ for gradient clipping during training.

The loss function used during training is a weighted sum of binary cross-entropy ($BCE$) and the logarithm of the soft Dice coefficient. We set $\alpha = 0.75$, $\beta=0.5$, and $\gamma = 0.25$ in equation \eqref{eq: pl_weights}. Additional information on our experimental section can be found in the supplementary materials. 

\section{Experimental Results}
\label{sec: Experimental Results}
Table~\ref{tab:IoU-Dice} compares the performance of all evaluated methods.
Accordingly, DeepPyramid, Unet++, and PSPNet+ are the top three segmentation methods in terms of IoU for the relevant objects in cataract surgery videos. However, DeepPyramid shows considerable improvements in segmentation accuracy compared to the second-best approach in each class. Specifically, DeepPyramid achieves more than $4\%$ improvement in lens segmentation ($85.61\%$ vs. $81.32\%$) and more than $4\%$ improvement in instrument segmentation ($74.40\%$ vs. $70.11\%$) compared to UNet++. Similarly, DeepPyramid achieves the highest dice coefficient compared to the evaluated approaches for all classes.

Table~\ref{tab:modules} validates the effectiveness of the proposed modules in an ablation study while also showing the impact on the different segmentation classes. The PVF module appears to enhance the performance for the cornea and instrument segmentation ($2.41\%$ and $2.76\%$ improvement in IoU, respectively). This improvement is most likely due to the ability of the PVF module to provide a global view of varying-size sub-regions centered around each spatial position. Such a global view can reinforce semantic representation in the regions corresponding to blunt edges and reflections. Due to scale variance in instruments, the DPR module boosts the segmentation performance for surgical instruments. The addition of the $P\mathcal{L}$ loss results in the improvement in IoU for all the relevant classes, especially the lens (roughly $2\%$ improvement) and instrument ($1.64\%$ improvement) classes. The combination of PVF, DPR, and $P\mathcal{L}$ show a marked $4.58\%$ improvement in instrument segmentation and $4.22\%$ improvement in cornea segmentation (based on IoU\%). These modules improve the IoU for the lens and pupil by $2.85\%$ and $1.43\%$, respectively. Overall, the addition of the different proposed components in DeepPyramid lead to considerable improvements in segmentation performance ($3.27\%$ improvement in IoU) when compared to the evaluated baselines.

\begin{table*}[t!]
\renewcommand{\arraystretch}{0.9}
\caption{Quantitative comparison between the proposed DeepPyramid and state-of-the-art approaches.}
\label{tab:IoU-Dice}
\centering
\begin{tabular}{l m{1.85cm}m{1.85cm}m{1.85cm}m{1.85cm}m{1.85cm}}
\specialrule{.12em}{.05em}{.05em}
&\multicolumn{5}{c}{IoU\%$\vert$Dice\%}\\\cmidrule(lr){2-6}
Network&Lens&Pupil&Cornea&Instrument&Mean\\\specialrule{.12em}{.05em}{.05em}
U-Net \cite{U-Net}& 58.19$\vert$67.91 & 85.51$\vert$89.36 & 79.83$\vert$86.20 & 56.12$\vert$67.02& 69.91$\vert$77.62\\
PSPNet+ \cite{PSPNet}& 80.56$\vert$88.89 & 93.23$\vert$96.45 &88.09$\vert$93.55 & 65.37$\vert$76.47& 81.81$\vert$88.84\\
UPerNet \cite{UPerNet}& 77.78$\vert$86.93 & 93.34$\vert$96.52 &86.62$\vert$92.67& 68.51$\vert$78.68& 81.56$\vert$88.70\\
FEDNet \cite{FED-Net}& 78.12$\vert$87.38 & 93.93$\vert$96.85 & 85.73$\vert$92.10 &65.13$\vert$76.11 & 80.72$\vert$88.11\\
RAUNet \cite{RAUNet}& 76.40$\vert$85.34 & 89.26$\vert$94.26 & 85.73$\vert$92.10 & 65.13$\vert$76.11 & 79.13$\vert$86.95\\
CE-Net \cite{CE-Net}& 68.40$\vert$80.43 & 83.59$\vert$90.89 & 83.47$\vert$90.85 & 61.57$\vert$74.64& 74.25$\vert$84.20\\
MultiResUNet \cite{MultiResUNet}
& 60.73$\vert$71.62 & 58.36$\vert$66.80 & 73.10$\vert$83.40 & 55.43$\vert$66.07 & 61.90$\vert$71.97\\
dU-Net \cite{dU-Net}& 59.83$\vert$69.46 & 71.86$\vert$79.53 & 82.39$\vert$90.00 & 61.36$\vert$71.55& 68.86$\vert$77.63\\
PAANet \cite{PAANet}& 74.92$\vert$84.83 & 90.02$\vert$94.59 & 86.75$\vert$92.71 & 64.47$\vert$75.24& 79.04$\vert$86.74\\
BARNet \cite{BARNet}& 67.33$\vert$78.85 & 91.33$\vert$95.32 & 83.98$\vert$91.09 & 66.72$\vert$77.14& 77.34$\vert$85.60\\
CPFNet \cite{CPFNet}& 73.56$\vert$83.74 &90.27$\vert$94.83 & 87.63$\vert$93.28 & 61.16$\vert$73.51 & 78.18$\vert$86.34\\
UNet++\slash DS \cite{UNet++}& 79.50$\vert$87.85 & 95.28$\vert$97.53 & 86.72$\vert$92.57 &66.05$\vert$75.91 & 81.88$\vert$88.46\\
UNet++ \cite{UNet++}& 81.32$\vert$89.34 & 95.66$\vert$97.77 & 85.08$\vert$91.72 & 70.11$\vert$79.56& 83.04$\vert$89.59\\\specialrule{.12em}{.05em}{.05em}
{\bf DeepPyramid}& \textbf{85.61} $\vert$ \textbf{91.98} & \textbf{96.56}$\vert$\textbf{98.24} & \textbf{90.24}$\vert$\textbf{94.63} & \textbf{74.40}$\vert$\textbf{83.30} & \textbf{86.70}$\vert$\textbf{92.03}\\\specialrule{.12em}{.05em}{.05em}
\end{tabular}
\label{tab:modules}
\end{table*}

\begin{figure}[!tb]
    \centering
    \includegraphics[width=0.99\columnwidth]{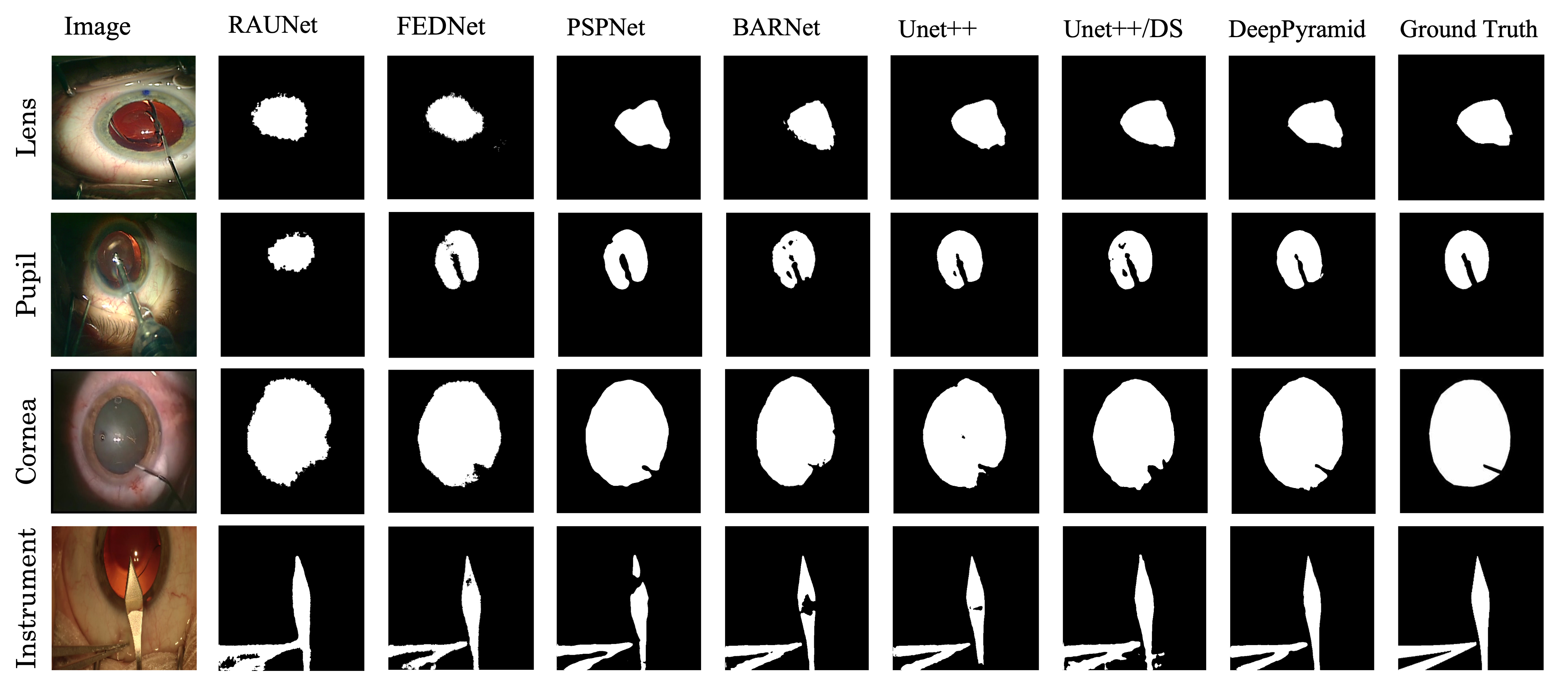}
    \caption{Qualitative comparisons among DeepPyramid and the six top-performing rival approaches for the relevant objects in cataract surgery videos. The representative images are selected from the test set.}
    \label{fig:Vis}
\end{figure}

\begin{table*}[t!]
\renewcommand{\arraystretch}{0.9}
\caption{Impact of different modules on DeepPyramid's performance (ablation study).}
\label{tab:ablation2}
\centering

\begin{tabular}{m{0.7cm} m{0.7cm} m{0.7cm} m{1.4cm}m{1.4cm}m{1.4cm}m{1.4cm}m{1.4cm}m{1.4cm}}
\specialrule{.12em}{.05em}{.05em}
\multicolumn{3}{c}{Modules}&&\multicolumn{5}{c}{IoU\%\slash Dice\%}\\\cmidrule(lr){1-3}\cmidrule(lr){5-9}
PVF&DPR&$P\mathcal{L}$&Params&Lens&Pupil&Cornea&Instrument&Overall\\\specialrule{.12em}{.05em}{.05em}
\xmark&\xmark&\xmark&22.55 M&82.98\slash90.44&95.13\slash97.48&86.02\slash92.28&69.82\slash79.05&83.49\slash89.81\\
\checkmark&\xmark&\xmark&22.99 M&83.73\slash90.79&96.04\slash97.95&88.43\slash93.77&72.58\slash81.84&85.19\slash91.09\\
\xmark&\checkmark&\xmark&23.17 M&81.85\slash89.58&95.32\slash97.59&86.43\slash92.55&71.57\slash80.60&83.79\slash90.08\\
\checkmark&\checkmark&\xmark&23.62 M&83.85\slash90.89&95.70\slash97.79&89.36\slash94.29&72.76\slash82.00&85.42\slash91.24\\
\checkmark&\checkmark&\checkmark&23.62 M&\textbf{85.84}\slash\textbf{91.98}&\textbf{96.56}\slash\textbf{98.24}&\textbf{90.24}\slash\textbf{94.77}&\textbf{74.40}\slash\textbf{83.30}&\textbf{86.76}\slash\textbf{92.07}\\ \specialrule{.12em}{.05em}{.05em}
\end{tabular}
\label{tab:modules}
\end{table*}
Fig.~\ref{fig:Vis} illustrates the qualitative results of our method and evaluated baselines. Specifically, we see the effectiveness DeepPyramid has in segmenting challenging cases. Taking advantage of the pyramid view provided by the PVF module, DeepPyramid can handle reflection and brightness variation in instruments, blunt edges in the cornea, color and texture variation in the pupil, as well as transparency in the lens. Furthermore, powered by deformable pyramid reception, DeepPyramid can tackle scale variations in instruments and blunt edges in the cornea. In particular, we see from Fig.~\ref{fig:Vis} that DeepPyramid shows much less distortion in the region of edges, especially in the case of the cornea. Furthermore, based on these qualitative experiments, DeepPyramid shows much better precision and recall in the narrow regions for segmenting the instruments and other relevant objects in the case of occlusion by the instruments. Further results are shown in the supplementary materials of the paper.

\section{Conclusion}
\label{sec: Conclusion}
 In this work, we have proposed a novel network architecture for semantic segmentation in cataract surgery videos. The proposed architecture takes advantage of two modules, namely ``Pyramid View Fusion" and ``Deformable Pyramid Reception", as well as a dedicated ``Pyramid Loss",  to simultaneously deal with (i) geometric transformations such as scale variation and deformability, (ii) blur degradation and blunt edges, and (iii) transparency, texture and color variation typically observed in cataract surgery images. We show in our experiments that our approach provides state-of-the-art performances in segmenting key anatomical structures and surgical instruments typical with such surgeries. Beyond this, we demonstrate that our approach outperforms a large number of recent segmentation methods by a considerable margin. The proposed architecture can also be adopted for various other medical image segmentation and general semantic segmentation problems.

\bibliographystyle{splncs04}
\bibliography{bibtex.bib}
\end{document}